\documentclass{article}
\usepackage{tikz}
\usepackage{pgfplots}
\usepackage{amsmath,epsfig}
\usepackage[preprint]{spconfa4}
\copyrightnotice{978-1-6654-3864-3/21/\$31.00 ©2021 IEEE}

\let\OLDthebibliography\thebibliography
\renewcommand\thebibliography[1]{
  \OLDthebibliography{#1}
  \setlength{\parskip}{0pt}
  \setlength{\itemsep}{0pt plus 0.3ex}
}

\usepackage{amssymb}
\usepackage{hyperref}
\usepackage{url}

\begin{document}\sloppy
\topmargin=0mm

\def\x{{\mathbf x}}
\def\L{{\cal L}}

\title{GSVNet: Guided Spatially-Varying Convolution for Fast Semantic Segmentation on Video}
%
\name{Shih-Po Lee, Si-Cun Chen, Wen-Hsiao Peng}
\address{National Yang Ming Chiao Tung University, Taiwan \\
\{splee, sicun.mapl.cs09, wpeng\}@nycu.edu.tw}

\maketitle

\newcommand\MakeLabel[1]{\phantomsubcaption\label{#1}(\subref{#1})}
\newcommand{\etal}{\textit{et al.}}
\definecolor{green2}{RGB}{31, 122, 0}
\definecolor{purple1}{RGB}{195,0,255}
\definecolor{purple2}{RGB}{255, 158, 247}
\definecolor{blue1}{RGB}{0, 199, 176}
\definecolor{mod}{RGB}{255,0,0}

\begin{abstract}
\textcolor{black}{This paper addresses fast semantic segmentation on video. Video segmentation often calls for real-time, or even faster than real-time, processing. One common recipe for conserving computation arising from feature extraction is to propagate features of few selected keyframes. However, recent advances in fast image segmentation make these solutions less attractive. To leverage fast image segmentation for furthering video segmentation, we propose a simple yet efficient propagation framework. Specifically, we perform lightweight flow estimation in 1/8-downscaled image space for temporal warping in segmentation outpace space. Moreover, we introduce a guided spatially-varying convolution for fusing segmentations derived from the previous and current frames, to mitigate propagation error and enable lightweight feature extraction on non-keyframes. Experimental results on Cityscapes and CamVid show that our scheme achieves the state-of-the-art accuracy-throughput trade-off on video segmentation.}
\end{abstract}


\vspace{-0.6cm}
\section{INTRODUCTION}
    \raggedbottom
    Video semantic segmentation is a compute-intensive vision task. It aims at classifying pixels in video frames into semantic classes. This task often has real-time or even faster than real-time requirements in data-center applications, in order to process hours-long videos in a much shorter time. The ever increasing video resolution in both spatial and temporal dimensions makes real-time processing even more challenging.
    
    A simple approach to video segmentation is to adopt image-based processing; that is, to process individual video frames independently using an off-the-shelf image segmentation network. This approach was once considered prohibitively expensive, given that most image segmentation networks, like DeepLabv3+ \cite{deeplabv3+}, usually optimized the feature extraction for segmentation accuracy rather than throughput. 
    
    
    To address the prolonged feature extraction, several fast video segmentation frameworks are proposed. One common recipe is to propagate features of few selected keyframes, in order to conserve computation for feature extraction on subsequent non-keyframes \cite{Clockwork, DFF}. This is motivated by the high correlation between consecutive video frames. Due to the evolution of video content in the temporal dimension, some \cite{Accel, LLVSS} additionally introduce lightweight feature extraction, followed by fusion of spatio-temporal features \cite{Accel, LLVSS}, for non-keyframes to cope with scene changes or dis-occlusion. Others propose adaptive keyframe selection \cite{DVSN, LLVSS} or bi-directional feature propagation \cite{jain2018fast} for alleviating error propagation. Meanwhile, Xu \etal \cite{DVSN} perform adaptive inference for cropped regions with scene changes.
    
    However, recent advances in fast image segmentation \cite{bisenet, swiftnet, fastscnn} make video-based solutions less attractive. For example, BiSeNet~\cite{bisenet} and SwiftNet~\cite{swiftnet} can now process high-definition ($2048 \times 1024$) videos at 40 to 60 frames per second on modern graphics processing units (GPU) while achieving reasonably good segmentation accuracy.
    A question that arises is whether video-based approaches can benefit from these advanced image segmentation networks. The answer relies crucially on how to address the following challenges. First, the increasing spatial resolution of videos may render optical flow estimation too expensive. Second, the excessive amount of channels in feature space may make the propagation of features time-consuming. Third, the feature extraction for non-keyframes has to be even more lightweight. Lastly, errors resulting from imperfect flow estimation and feature extraction may be propagated along the temporal dimension. 

    To tackle these issues, we propose a simple yet efficient propagation framework, termed GSVNet, for fast semantic segmentation on video. Our contributions include: (1) to conserve computation for temporal propagation, we perform lightweight flow estimation in 1/8-downscaled image space for warping in segmentation outpace space; and (2) to mitigate propagation error and enable lightweight feature extraction on non-keyframes, we introduce a guided spatially-varying convolution for fusing segmentations derived from the previous and current frames.
    
    Experimental results on Cityscapes~\cite{cityscapes} show when working with BiSeNet~\cite{bisenet} and SwiftNet~\cite{swiftnet}, our scheme can process up to 142 high-definition video frames per second on GTX 1080Ti with $71.8\%$ accuracy in terms of Mean Intersection over Union (mIoU). It achieves the state-of-the-art accuracy-throughput trade-off on video segmentation and can work with any off-the-shelf image segmentation network.
\vspace{-0.3cm}
\section{RELATED WORK}
\vspace{-0.2cm}
\label{sec:related}
\noindent \textbf{Image Semantic Segmentation:} Image semantic segmentation models \cite{deeplabv3+, pspnet} have achieved great success in segmentation accuracy by incorporating sophisticated feature extractors and task decoders \cite{deeplabv3+, pspnet}. To achieve better accuracy-throughput trade-offs, recent research \cite{espnet, bisenet, fastscnn} has been focused on making feature extractors lightweight and less sensitive to the adaptation of input resolution. To this end, 
Yu \etal ~\cite{bisenet} introduce a cost-effective feature extractor by including a spatial path for preserving spatial details and a context path for capturing contextual information in a wide receptive field.
In another attempt,  Orsic \etal ~\cite{swiftnet} take the advantage of transfer learning by using pre-trained encoder on ImageNet and adopting a simple upsampling decoder with lateral connections.

\noindent \textbf{Video Semantic Segmentation:} Efficient video segmentation is another active research area. Unlike images, consecutive video frames usually have a high correlation or similarity. To conserve computation for feature extraction, several works leverage the temporal correlation between video frames to reuse features of selected keyframes on non-keyframes. Shelhamer \etal~\cite{Clockwork} employ features at different stages of the network from previous frames. For feature propagation, Zhu \etal~ \cite{DFF} use optical flow estimated by a flow network, while Li \etal~\cite{LLVSS} adopt spatially variant convolution. \cite{Accel, LLVSS} additionally introduce lightweight feature extraction for non-keyframes together with spatio-temporal feature fusion, to reduce error propagation arising from scene changes or warping errors. 
However, the emergence of lightweight image segmentation calls for a careful rethink of these strategies. 
Specifically, the feature extraction on non-keyframes must be even more lightweight and the extracted features must be used to their fullest potential to mitigate error propagation. 


\begin{figure}
    \centering
    \includegraphics[width=\linewidth]{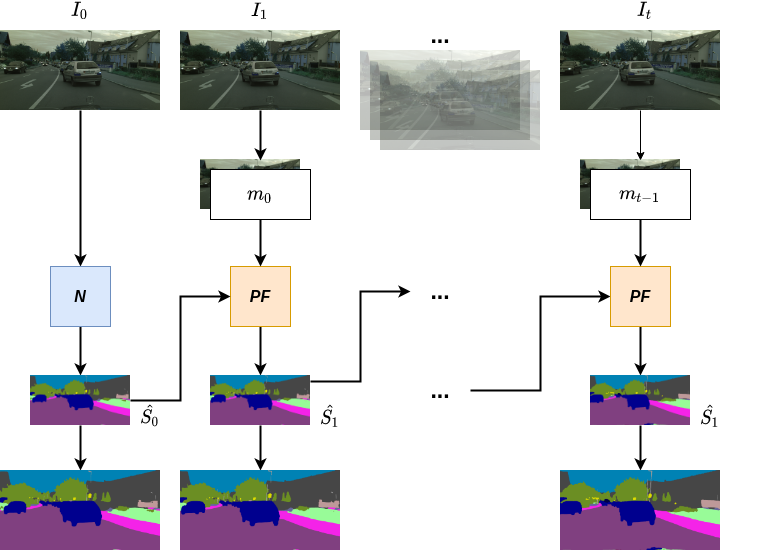}
    \caption{Illustration of our proposed propagation framework, where $m_i$ is the optical flow map, $N$ denotes the image-based segmentation network and $\mathcal{PF}$ indicates our propagation network.}
    \label{fig:system}
    \vspace{-0.3cm}
\end{figure}
\begin{figure*}[t]
    \centering
    \includegraphics[width=1.\linewidth]{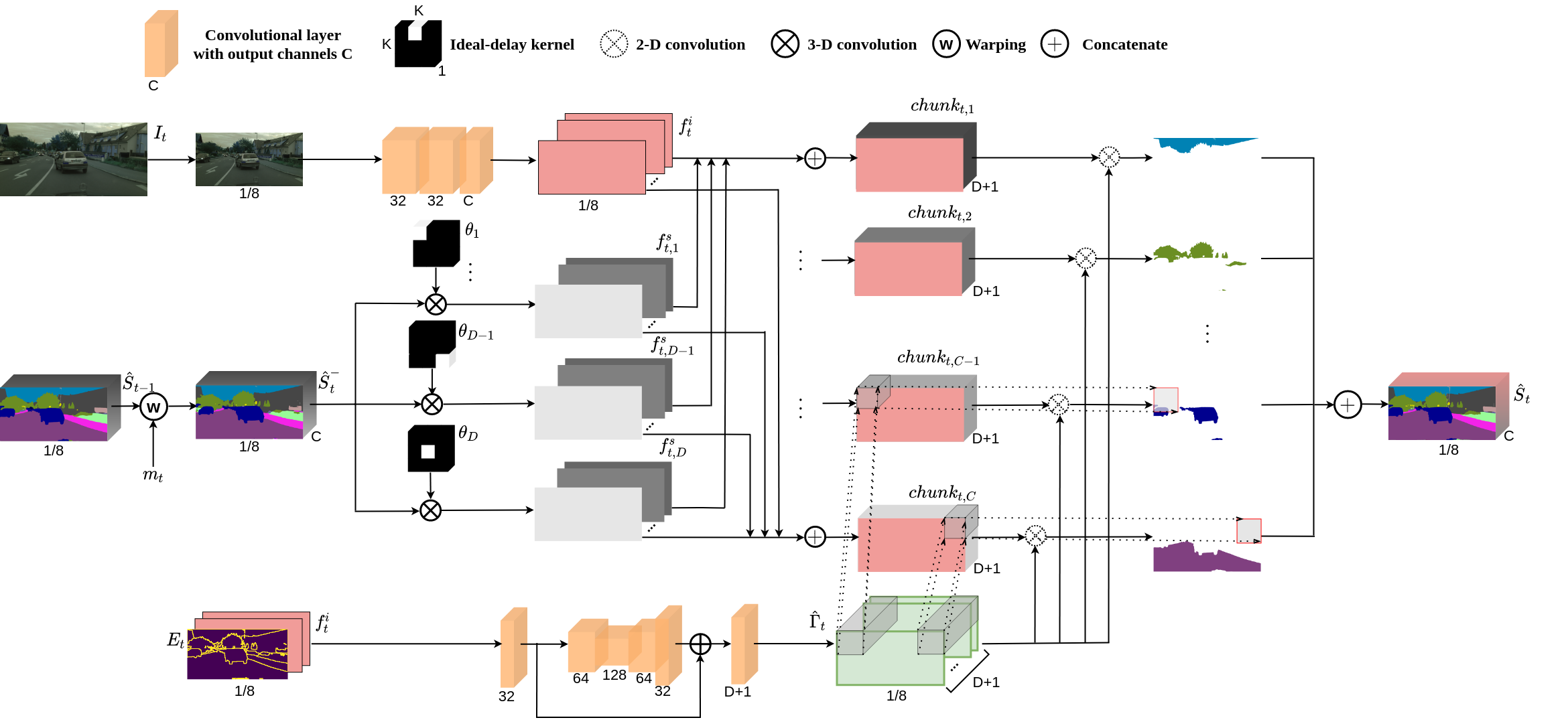}
    \caption{Architecture diagram of the proposed propagation network $\mathcal{PF}$.}
    \label{fig:arc}
\end{figure*}
\vspace{-0.3cm}
\section{METHOD}
\label{sec:method}
This work addresses the problem of efficient semantic segmentation on video. Given an input video consisting of a set 
$\{I_t\}_{t=0}^{N-1}$ of video frames, 
each being of dimension $3 \times H \times W$, our task is to predict for every video frame 
$I_{t}$ its downscaled, 
semantic segmentation $\hat{S}_t \in \mathbb{R}^{C \times H/8 \times W/8}$ 
with $C$ classes, aiming to strike a good balance between \textit{accuracy} 
and \textit{throughput}. 
Following common practice, the final 
segmentation is obtained by upsampling 
$\hat{S}_t$ to the full resolution, 
where the segmentation accuracy is measured.
\vspace{-0.3cm}
\subsection{System Overview}
\vspace{-0.2cm}
Fig.~\ref{fig:system} illustrates our proposed method for efficient semantic segmentation on video. The process begins by segmenting the first keyframe to obtain its semantic segmentation $\hat{S}_0$ with an image-based segmentation network $\mathcal{N}$. Our propagation framework $\mathcal{PF}$ then propagates temporally the segmentation $\hat{S}_0$ (respectively, $\hat{S}_{t-1}$) of the previous step to assist in predicting the segmentation $\hat{S}_1$ (respectively, $\hat{S}_{t}$) for the current non-keyframe $I_1$ (respectively, $I_{t}$). 
The process repeats until the next keyframe is reached. 


\vspace{-0.3cm}
\subsection{Spatiotemporal Propagation Framework}

Fig.~\ref{fig:arc} shows the pipeline of our segmentation propagation framework $\mathcal{PF}$. Starting with the downscaled segmentation $\hat{S}_{t-1} \in \mathbb{R}^{C \times H/8 \times W/8}$ for the previous frame $I_{t-1}$, we (1) perform temporal (backward) warping of $\hat{S}_{t-1}$ based on the optical flow $m_{t} \in \mathbb{R}^{2 \times H/8 \times W/8}$ to arrive at an initial estimate $\hat{S}_t^- \in \mathbb{R}^{C \times H/8 \times W/8}$ of the downscaled segmentation for the current frame $I_t$. 
It is obvious that not every semantic prediction in $\hat{S}_t^-$ is reliable.
The errors can propagate along the temporal dimension through the warping process. Moreover, the optical flow $m_t$ may be error-prone due to dis-occlusion or imperfect flow estimation. We thus (2) refine the warped segmentation $\hat{S}_t^-$ by applying a \textit{guided spatially-varying} convolution. This is achieved in three sequential steps. The first convolves the $\hat{S}_t^-$ with several \textit{ideal-delay} kernels $\theta_1,\theta_2,\ldots,\theta_D$, yielding $\{f_{t,d}^s\}_{d=1}^{D},f_{t,d}^s \in \mathbb{R}^{C \times H/8 \times W/8}$. The second arranges $\{f_{t,d}^s\}_{d=1}^D$ together with a crude estimate $f_t^i \in \mathbb{R}^{C \times H/8 \times W/8}$ of the current frame's segmentation from a lightweight network into $C$ chunks, each concatenating the channels of the same index from $\{f_{t,d}^s\}_{d=1}^D$ and $f_t^i$. The last applies a spatially-varying $1 \times 1$ convolution across the channels of each of these $C$ chunks, with their results concatenated together to form our downscaled estimate $\hat{S}_t \in \mathbb{R}^{C \times H/8 \times W/8}$ of the current frame's segmentation.
 


\vspace{-0.3cm}
\subsection{Temporal Segmentation Warping}\label{ssec:segwarp}
This very first step performs temporal warping of $\hat{S}_{t-1}$ to arrive at an initial estimate $\hat{S}_t^-$ of the current frame's segmentation. In symbols, we have
\begin{equation}
\label{eq:warp}
    \hat{S}_{t}^-(c,x,y)  = \hat{S}_{t-1}(c,x+m_t^{(x)}(x,y),y+m_t^{(y)}(x,y)), \forall c \in C
    \vspace{0.25em}
\end{equation}
where $(x,y),1 \leq x \leq W/8, 1 \leq y \leq H/8$ denotes pixel locations in the downscaled segmentation and $m_t=(m_t^{(x)},m_t^{(y)})$ is the optical flow describing the backward motion from $I_{t}$ to $I_{t-1}$ also in the downscaled domain, and is estimated by a lightweight optical flow network $O$ based on the hierarchical feature fusion block.
More details of this flow network can be found in the supplementary document.
Performing temporal warping in the 1/8-downscaled domain is motivated by two observations. First, lowering the spatial resolution can significantly reduce the runtime costs of both optical flow estimation and temporal warping. Second, the 1/8 downscaling factor is chosen empirically to match common practice in most segmentation network design that the final high-resolution segmentation is usually recovered by interpolating the network output by a factor of 8.

\vspace{-0.3cm}
\subsection{Guided Spatially-Varying Convolution}\label{subsec:gsvc}

To correct unreliable segmentation predictions in  $\hat{S}_t^-$, we resort to a guided spatially-varying filtering. This operation serves as a means to propagate \textit{spatially} within $\hat{S}_t^-$ the segmentation predictions from reliable regions to unreliable ones. Apparently, in what direction the propagation should be performed to correct the erroneous prediction at a pixel depends on the error pattern in its local neighborhood. As such, our filtering kernel is designed to be both context- and pixel-adaptive.

\vspace{-0.3cm}
\subsubsection{Spatial Propagation with Ideal-delay Kernels}
\vspace{-0.2cm}
Such spatial propagation starts by convolving the temporally warped segmentation $\hat{S}_t^-$ with several ideal-delay kernels $\theta_1, \theta_2, \ldots, \theta_D$. 
As a result of the convolution, each of these kernels shifts the $\hat{S}_t^-$ channel-wise by one pixel or more in a certain direction. When viewed from each pixel's perspective, these shifting operations line up its surrounding semantic predictions as potential candidates for propagation. For implementation, we consider each ideal-delay kernel as a 3-D tensor of size $1 \times K \times K$, where $K$ indicates the propagation range measured in the maximum horizontal or vertical displacement of the unit impulse from the origin. It is then convolved with $\hat{S}_t^-$, also regarded as a 3-D tensor, using 3-D convolution $\otimes$ to get $f_{t,d}^s = \hat{S}_{t}^- \otimes \theta_{d}$, where $f_{t,d}^s \in \mathbb{R}^{C \times H/8 \times W/8}$, $d=1,2,\ldots,D$.

\vspace{-0.3cm}
\subsubsection{Lightweight Intra-frame Segmentation}
\vspace{-0.2cm}
Recognizing that there may be areas in a video frame where new image contents emerge or our spatial propagation may not work well, we further develop a lightweight network $\varphi$ for segmentation in these areas. The former is also known as the dis-occluded areas; the latter is because our spatial propagation is susceptible to the errors of the previous frame's segmentation and the optical flow estimation. The module is showed in the upper branch of Fig.~\ref{fig:arc}. Since these areas normally represent only a small percentage of pixels in a video frame, we design $\varphi$ for low complexity, with its prediction accuracy focused on those error-prone areas. Our current implementation comprises only 3 convolutional layers with kernel size $3 \times 3$ without resolution downscaling. 
As shown in Fig.~\ref{fig:arc}, it takes as input the 1/8-downscaled current video frame and outputs the segmentation $f_t^i = \varphi(\hat{S}_{t}^-), f_t^i \in \mathbb{R}^{C \times H/8 \times W/8}$.




\vspace{-0.3cm}
\subsubsection{Guided Dynamic Filtering}
\vspace{-0.2cm}
Given $\{f_{t,d}^s\}_{d=1}^D$ and $f_t^i$ from the previous two steps, the guided dynamic filtering forms a semantic prediction of every pixel by spatially-varying $1 \times 1$ filtering. To proceed, we first organize $\{f_{t,d}^s\}_{d=1}^D$ and $f_t^i$ into $C$ chunks. Each chunk concatenates the channels of the same semantic class from $\{f_{t,d}^s\}_{d=1}^D$ and $f_t^i$; that is, the resulting channels of a chunk correspond to semantic predictions of the same class yet shifted spatially from $\hat{S}_t^-$ in different directions. We then use a guiding network $G$ to produce the $1 \times 1$ kernel for every pixel. In particular, this kernel varies from one pixel location to another, determining dynamically how the spatial propagation is conducted at every pixel location or whether the intra-frame segmentation should be weighted more heavily.

The process is illustrated in the lower and right parts of Fig.~\ref{fig:arc}. For implementation, the guiding network $\Gamma_t = G(f_t^i, E_t)$ takes as inputs $f_t^i$ and $E_t$, where $E_t = \sigma(\hat{S}_{t}^- \otimes M)$ is an edge map derived from convolving $\hat{S}_{t}^-$ with a Laplacian kernel $M$. In particular, before the convolution, we take the argmax operation with respect to $\hat{S}_{t}^-$ to keep only the structure information inherent in the segmentation map.
The output of the convolution is further passed through a Sigmoid function $\sigma$ to clip the edge response.   
This design is inspired by the observation that $\hat{S}_{t}^-$ tends to be erroneous at the object boundaries and in regions where the structure contour of $\hat{S}_{t}^-$ is inconsistent with that of $f_t^i$, which captures rough semantic predictions. Finally, the network output $\Gamma_t \in \mathbb{R}^{(D+1) \times H/8 \times W/8}$ is the $1 \times 1$ kernel for every pixel location $(x,y),1 \leq x \leq W/8, 1 \leq y \leq H/8$. The filtering of every chunk then follows by evaluating     
\begin{equation}
\label{eq:weighed_sum}
    \hat{S}_{t,c}(x,y) = 
    \hat{\Gamma}_{t}(x,y) \odot chunk_{t,c}(x,y),\forall c,x,y 
    \vspace{0.25em}
\end{equation}
where $\hat{\Gamma}_{t}(x,y)$ denotes the normalized kernel at $(x,y)$ that has a unity gain across channels and $\odot$ is the inner product. The filtered results, when pooled together, form the final segmentation output $\hat{S}_t$. 


\textcolor{black}{
We train the propagation framework $\mathcal{PF}$ end-to-end. The training objective involves an ordinary cross-entropy loss 
imposed on the final segmentation output.}


\section{EXPERIMENTS}
\label{sec:experiments}
    \subsection{Setup}\label{ssec:exp_implementation}
        \noindent \textbf{Datasets:} \textcolor{black}{We validate our method on Cityscapes \cite{cityscapes} and CamVid \cite{camvid}, the datasets for scene understanding. Cityscapes includes video snippets of urban scenes, of which 2975/500/1525 are for training/validation/test. Each snippet of resolution $2048 \times 1024$ is 30 frames long, with only the 20th frame annotated. CamVid has 4 video sequences. In each video, only one frame in every 30 frames is annotated, leading to a total of 701 labeled video frames of size $960 \times 720$. Among these video frames, 367/101/233 are for training/validation/test.}

        \noindent \textbf{Accuracy Metrics:} For evaluating the segmentation accuracy, we follow the protocol in \cite{DFF, Accel} to measure the Mean Intersection over Union (mIoU) over $l$ image pairs $(20 - i, 20)$ in every snippet with $i = 0, 1,\ldots,l-1$. For each image pair $(a,b)$, we propagate the segmentation for frame $a$ (keyframe) recursively to frame $b$ (non-keyframe). The mIoU measured this way is known as the \textit{average} mIoU. \textcolor{black}{We also report the \textit{minimum} mIoU measured with respect to the non-keyframe farthest away from the keyframe in every snippet, i.e. the non-keyframe that is expected to be impacted the most by error propagation in the temporal dimension.} 
        Unless otherwise stated, the keyframe interval $l$ is defaulted to 5.
        \textcolor{black}{Note that all the mIoU numbers reported are measured at full resolution.}
        
        \noindent \textbf{Complexity Metrics:} For complexity assessment, we report the throughput in \textit{frames per second} (FPS) on GTX 1080Ti, the number of network parameters in bytes, and the average number of \textit{floating-point operations per second} (FLOPS) over a keyframe interval. 
        
        \noindent \textbf{Implementation Details:} \textcolor{black}{We implement our network on Pytorch, and use pre-trained SwiftNet-R18~\cite{swiftnet}, termed SN-R18 or BiSeNet-R18, termed BN-R18, as the segmentation model for keyframes. Both SN-R18 and BN-R18 adopt the ResNet-18 backbone. 
        At inference time, the size of the keyframe is downscaled for the best accuracy-throughput trade-off. By the same token, our propagation module operates at one-eighth the resolution of the input video frame, followed by an interpolation of the segmentation output to the full resolution for mIoU measurement.
        Similar to \cite{DFF, Accel}, we train our network for keyframe intervals ranging from 1 to 5. 
        We use stochastic gradient descent (SGD) with momentum 0.9 and a learning rate of 0.002, which is decreased by a factor of 0.992 every 100 iterations. We set the batch size to 8 and the weight decay to 0.0005. }
        \textcolor{black}{Our model, exclusive of the pre-trained SwiftNet-R18 or BiSeNet-R18 for keyframe segmentation, is then trained end-to-end over image patches cropped from the input image. Each patch is three quarters of the input size. Our code and pre-trained weights are available online. \footnote{\url{https://github.com/robert80203/GSVNet}}}
    \newif\ifcomment

    
    \subsection{Results}
        
        
        \noindent \textbf{Accuracy-throughput Trade-off:}
        \textcolor{black}{Table~\ref{table:compare} compares the accuracy-throughput trade-offs of the competing methods.
        As shown, at FPS around 130, Ours-BN-R18~($l=4$) outperforms BiSeNet-R18 with input size 0.5 by 2.3\% mIoU. Likewise, Ours-SN-R18~($l=4$) surpasses SwiftNet-R18 with input size 0.5 by 3.1\% mIoU. Comparing with the video-based methods like \cite{Clockwork, Accel, DVSN}, our scheme achieves much higher FPS and mIoU. In particular, Ours-SN-R18~($l=2$) outpaces considerably \cite{LLVSS, td_net}, which target also high throughput, in FPS at the cost of a modest drop in mIoU. Results on Camvid~\cite{camvid} (Table~\ref{table:camvid}) show that our method runs faster than the image-based schemes~\cite{bisenet} while achieving higher or comparable mIoU. It is to be noted that the other video-based schemes can hardly compete with ours in FPS, although \cite{td_net} has higher mIoU due to the use of better backbones.} 
    \noindent \textbf{Network Parameters and FLOPS:}
    Table~\ref{table:complex} compares the number of network parameters and FLOPS. As can be seen, our scheme introduces 1.6M additional network parameters for segmenting non-keyframes, representing less than $3.3\%$ additional overhead relative to image-based segmentation with SwiftNet-R18 or BiSeNet-R18. It achieves the least number of FLOPS, with a FPS of 142 and 71.8$\%$ average mIoU. The FLOPS of our scheme is variable depending on the keyframe interval. On average, a non-keyframe requires 2.8G FLOPS, as compared to 58.5G FLOPS for processing a keyframe with SwiftNet-R18 (0.75). As such, the higher the keyframe interval, the lower the FLOPS. 
            \begin{table}[t]
        \centering
        \setlength\tabcolsep{0.3cm}
        \vspace{-0.5cm}
        
        \begin{tabular}{l|ccc}
            \hline
            Method &  Input & Avg. & FPS \\
             & size & mIoU & \\ 
            \hline
            CLK~\cite{Clockwork} & 1.0 & 64.4 & 6.3 \\
            LVS-LLS~\cite{LLVSS}  & 1.0 & 75.9 & 8.4 \\
            Accel-R18~\cite{Accel}  & 1.0 & 72.1 & 2.2 \\
            DVSNet~\cite{DVSN}  & 1.0 & 63.2 & 30.3 \\
            \textcolor{black}{TD-BN-R18}~\cite{td_net}  & 1.0 & 75.0 & 47 \\
            \hline
            Fast-SCNN~\cite{fastscnn} & 0.5 & 68.6 & 123.5 \\
            BiSeNet-R18~\cite{bisenet}  & 0.75 & 73.7 & 62 \\
            BiSeNet-R18~\cite{bisenet}  & 0.5 & 68.9 & 131 \\
            SwiftNet-R18~\cite{swiftnet}  & 0.75 & 74.4 & 63.2 \\
            SwiftNet-R18~\cite{swiftnet}  & 0.5 & 68.7 & 134 \\
            \hline
            Ours-BN-R18 (\textit{l}=2) & 0.75 & 72.9 & 98.5 \\
            Ours-BN-R18 (\textit{l}=3) & 0.75 & 72.0 & 123.4 \\
            Ours-BN-R18 (\textit{l}=4) & 0.75 & 71.2 & 140 \\
            Ours-SN-R18 (\textit{l}=2) & 0.75 & 73.6 & 100 \\
            Ours-SN-R18 (\textit{l}=3) & 0.75 & 72.6 & 125 \\
            Ours-SN-R18 (\textit{l}=4) & 0.75 & 71.8 & 142 \\
          \hline
          
        \end{tabular}
        
        \caption{\textcolor{black}{Accuracy and throughput comparison on Cityscapes dataset. The \textit{l} specifies the keyframe interval. The input size indicates the downscaling factor for keyframes in the video-based schemes or for input images in the image-based schemes.}}
        \label{table:compare}
    \end{table}
            \begin{table}[t]
        \centering
        \setlength\tabcolsep{0.1cm}
        
        \vspace{-0.2cm}
        \begin{tabular}{l|ccccc}
            \hline
            Method & Input & FPS & Key & Non-key & FLOPS\\ 
            & size &  & params & params & \\ 
            \hline
            
            BiSeNet-R18 & 0.75 & 62.0 & 49.0M & - & 58.0G \\
            SwiftNet-R18 & 0.75 & 63.2 & 47.2M & - & 58.5G \\
            SwiftNet-R18 & 0.5 & 134 & 47.2M & - & 26.0G \\ \hline
            
            
            Ours-SN-R18 (\textit{l}=4) & 0.75 & 142 & 47.2M  & 1.6M & 16.7G \\ 
          \hline
          
        \end{tabular}
        \caption{\textcolor{black}{Complexity comparison in terms of network parameters, FLOPS and FPS.}}
        \label{table:complex}
        \vspace{-0.4cm}
    \end{table}
        \begin{table}[t]
    \centering
    \setlength\tabcolsep{0.25em}
    \vspace{-0.5cm}
    \begin{tabular}{l|ccc}
        \hline
        Method & Input & Avg. & FPS\\ 
        & size & mIoU & \\
        \hline
        Accel~\cite{Accel} & 1.0 & 66.7 & 7.58\\
        TD-PSP18~\cite{td_net} & 1.0 & 71.0 & 25\\
        TD-PSP50~\cite{td_net} & 1.0 & 74.7 & 10\\
        \hline
        BiSeNet-R18~\cite{bisenet} & 0.75 & 66.6 & 142 \\
        BiSeNet-R18~\cite{bisenet} & 0.5 & 60.7 & 232\\ 
        \hline
        Ours-BN-R18 (\textit{l}=2) & 0.75 & 65.9 & 210  \\
        Ours-BN-R18 (\textit{l}=3) & 0.75 & 64.8 & 250  \\
      \hline
      
    \end{tabular}
    \caption{Accuracy and FPS comparison on CamVid test set.}
    \label{table:camvid}
\end{table}

    
    
           
    
    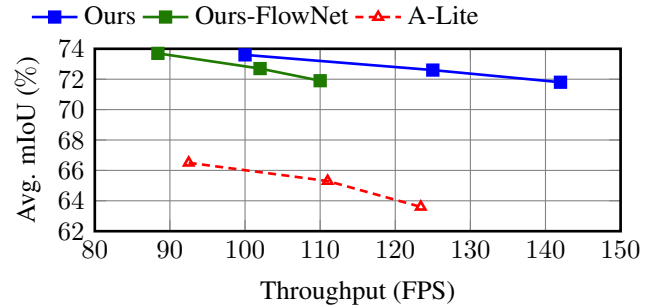
\begin{figure}[t]
        \setlength{\tabcolsep}{6pt}
        \begin{tikzpicture}
            \begin{axis}[
                title={},
                height=4cm,width=8.5cm,
                xlabel={Throughput (FPS)},
                ylabel={Avg. mIoU ($\%$)},
                ylabel style={at={(axis description cs:0.05,0.5)}},
                xlabel style={at={(axis description cs:0.5,0.0)}},
                xmin=80, xmax=150,
                ymin=62, ymax=74,
                xtick={80, 90, 100, 110, 120, 130, 140,150},
                ytick={62,64, 66, 68, 70, 72, 74},
                legend style={at={(-0.14, 1.3)},anchor=north west, legend  columns =4, draw=none},
                line width=1.0pt,
                mark size=2.0pt,
                xmajorgrids,
                ymajorgrids,
                legend cell align={left},
                grid style={solid, gray},
            ]
            \addplot[
                color=blue,
                mark=square*,
                ]
                coordinates {(100,73.6)(125,72.6)(142,71.8)};
            \addlegendentry{Ours}
            \addplot[
                color=green2,
                mark=square*,
                ]
                coordinates {(88.4,73.7)(102,72.7)(110,71.9)};
            \addlegendentry{Ours-FlowNet}
            
            
            \addplot[
                color=red,
                densely dashed,
                mark=triangle,
                every mark/.append style={solid},
                ]
                coordinates {(92.5,66.5)(111,65.3)(123.4,63.6)};
            \addlegendentry{A-Lite}
            \end{axis}
        \end{tikzpicture}
        \vspace{-0.5em}
        \vspace{-0.3cm}
        \caption{Comparison of accuracy-throughput curves. A-Lite adopts the same flow network and feature extractor as ours. Ours-FlowNet uses FlowNet2s for flow estimation.}
        \label{fig:ours_ax18_com}
    \end{figure}
    \begin{table}[t]
    \centering
    
    \vspace{-0.3cm}
    \begin{tabular}{p{0.1\linewidth}cccccc}
        \hline
         & & \multicolumn{5}{c}{Keyframe interval} \\ \cline{3-7}
        Metric & Intra Feature & 1 & 2 & 3 & 4 & 5 \\ \hline
        Min. & with & 74.4 & 72.9 & 70.7 & 69.6 & 67.7 \\ 
        mIoU & without & 74.4 & 72.1 & 69.1 & 67.4 & 65.1 \\
        \hline
    \end{tabular}
    \caption{Accuracy comparison between models with and without the use of the current frame's features (intra features). The accuracy is measured in minimum mIoU over a keyframe interval to emphasize the worst case.}
    \label{table:intra_com}
\end{table}
    \begin{figure}[ht]
            \setlength{\belowcaptionskip}{1pt}
            \vspace{-0.2cm}
            \centering
            \setlength\tabcolsep{0.1em}
            \begin{tabular}{@{}cccc@{}}
                \raisebox{1.5\normalbaselineskip}[0pt][0pt]{\rotatebox[origin=c]{90}{GT}} & & & \includegraphics[width=0.3\linewidth]{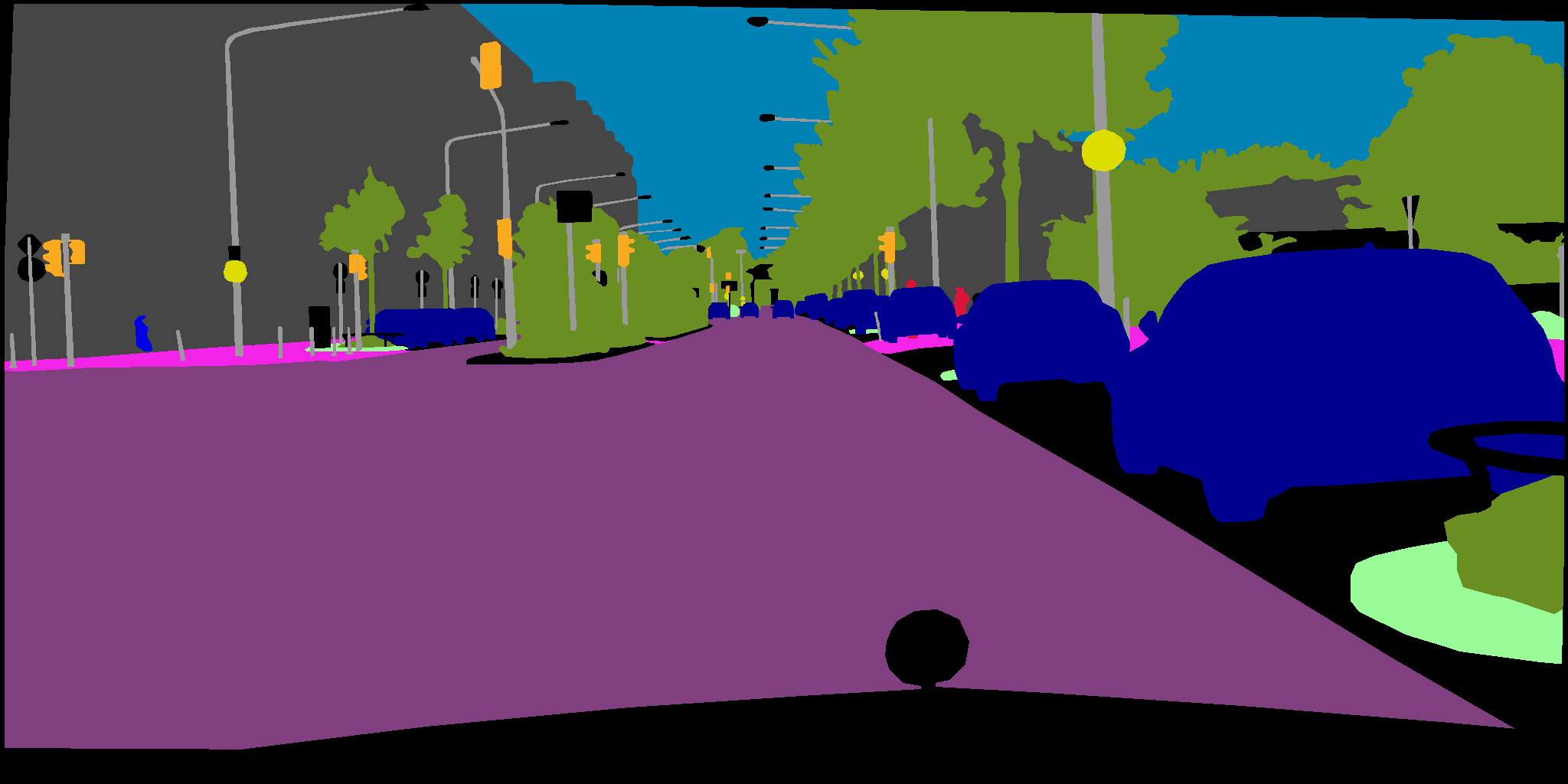} \\ \raisebox{1.5\normalbaselineskip}[0pt][0pt]{\rotatebox[origin=c]{90}{(a)}} &
                \includegraphics[width=0.3\linewidth]{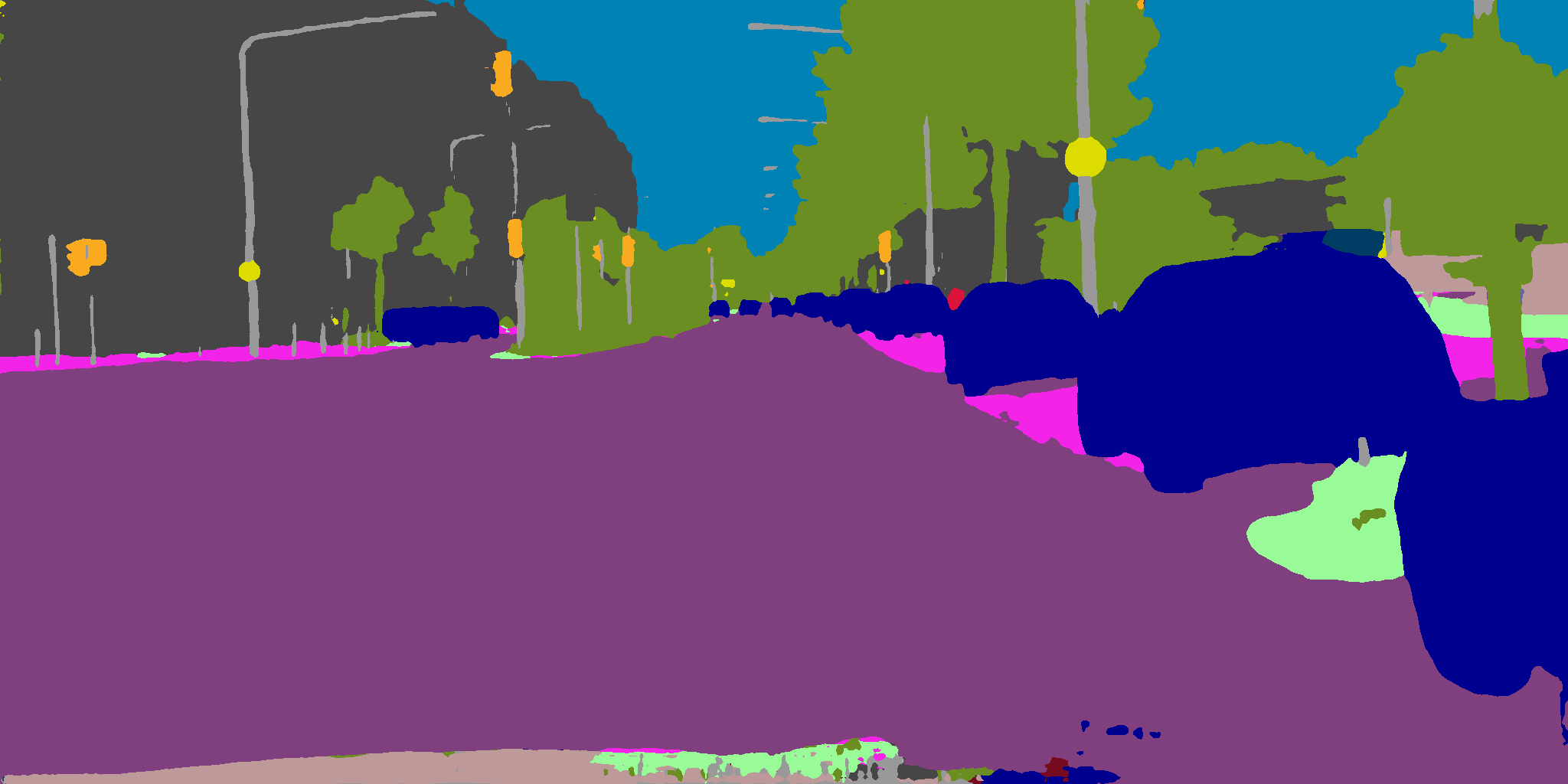} &
                
                \includegraphics[width=0.3\linewidth]{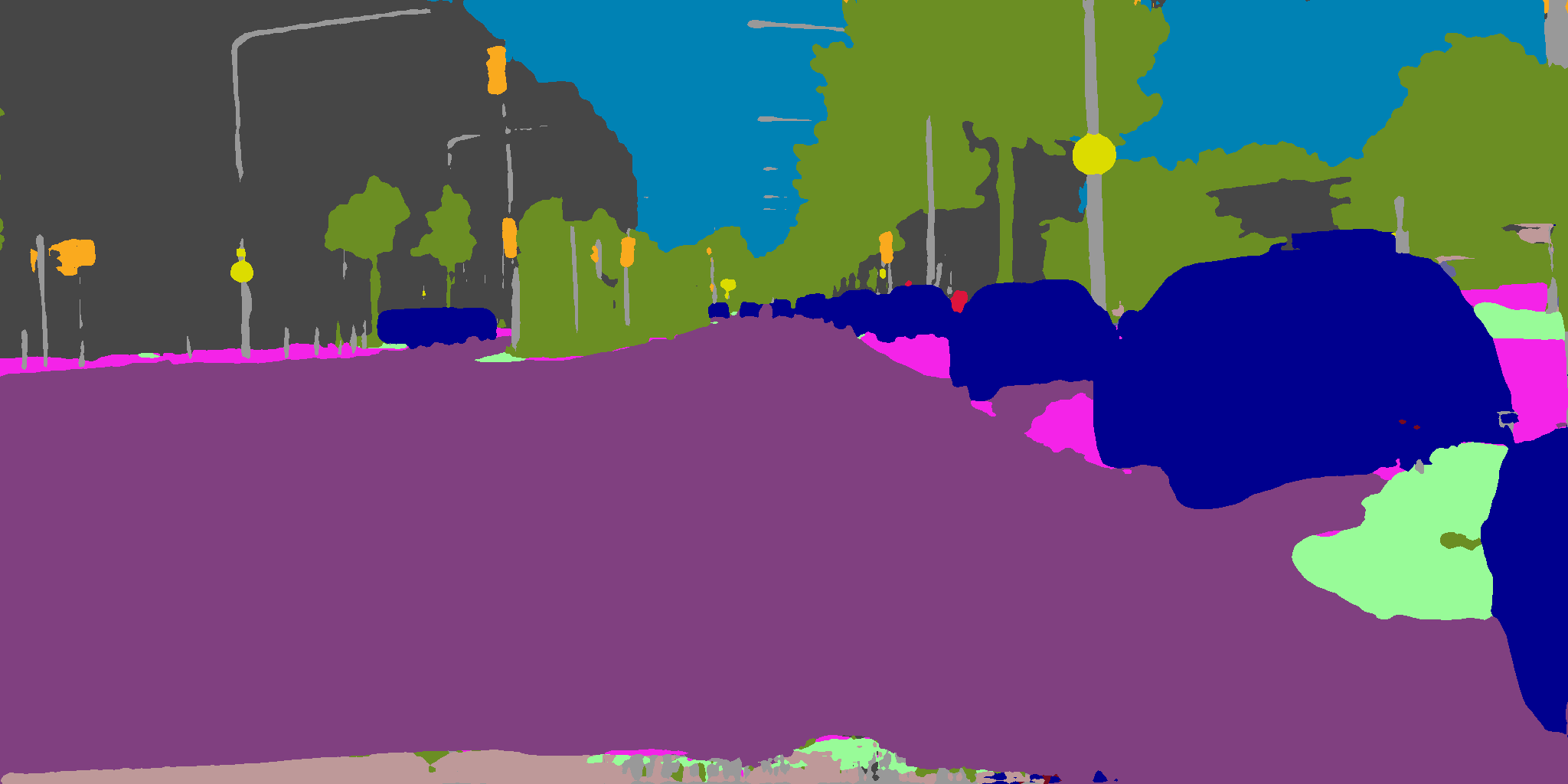} &
                
                \includegraphics[width=0.3\linewidth]{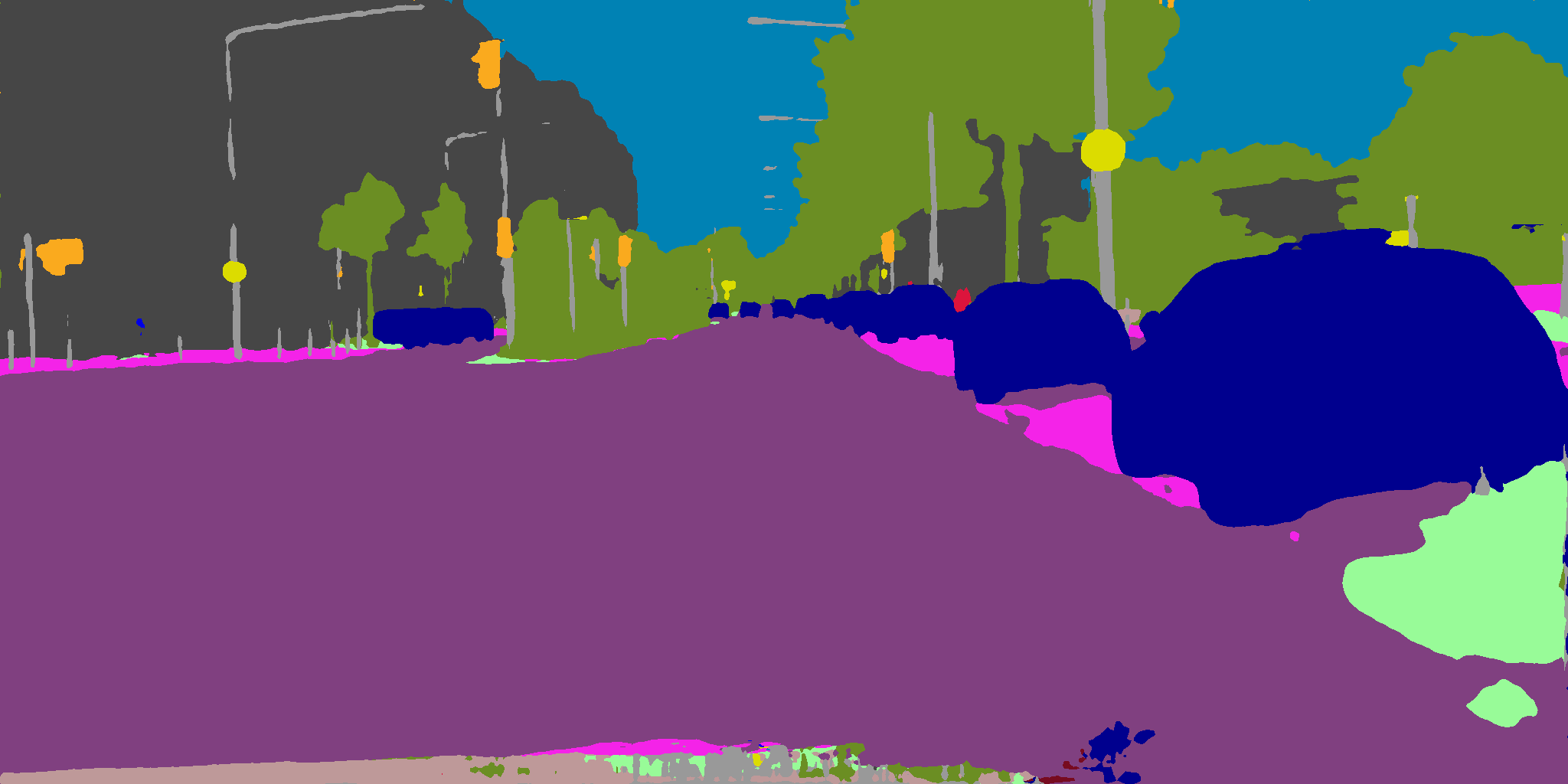} 
                
                \\
                
                \raisebox{1.5\normalbaselineskip}[0pt][0pt]{\rotatebox[origin=c]{90}{(b)}} & 
                \includegraphics[width=0.3\linewidth]{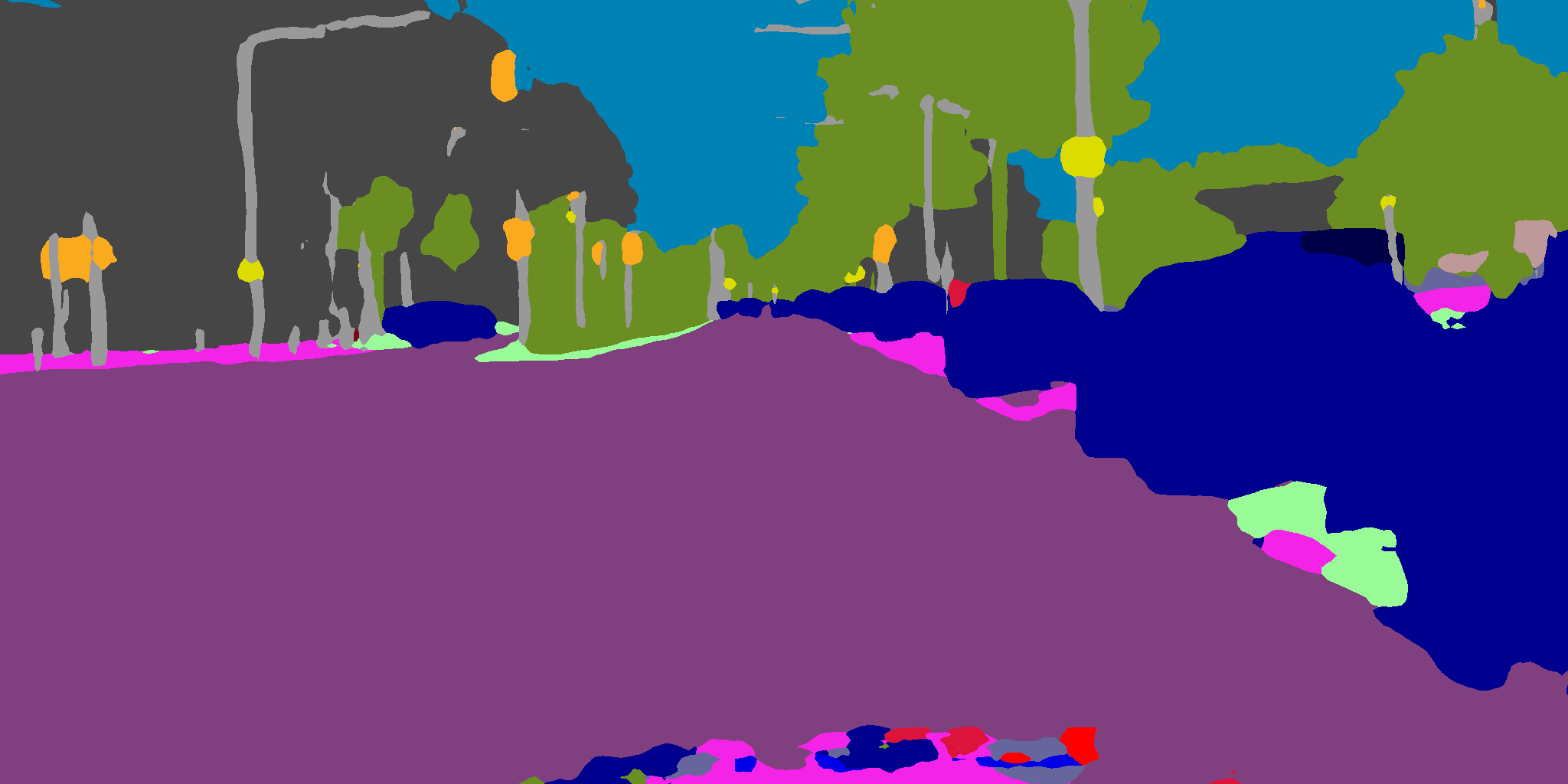} &
                \includegraphics[width=0.3\linewidth]{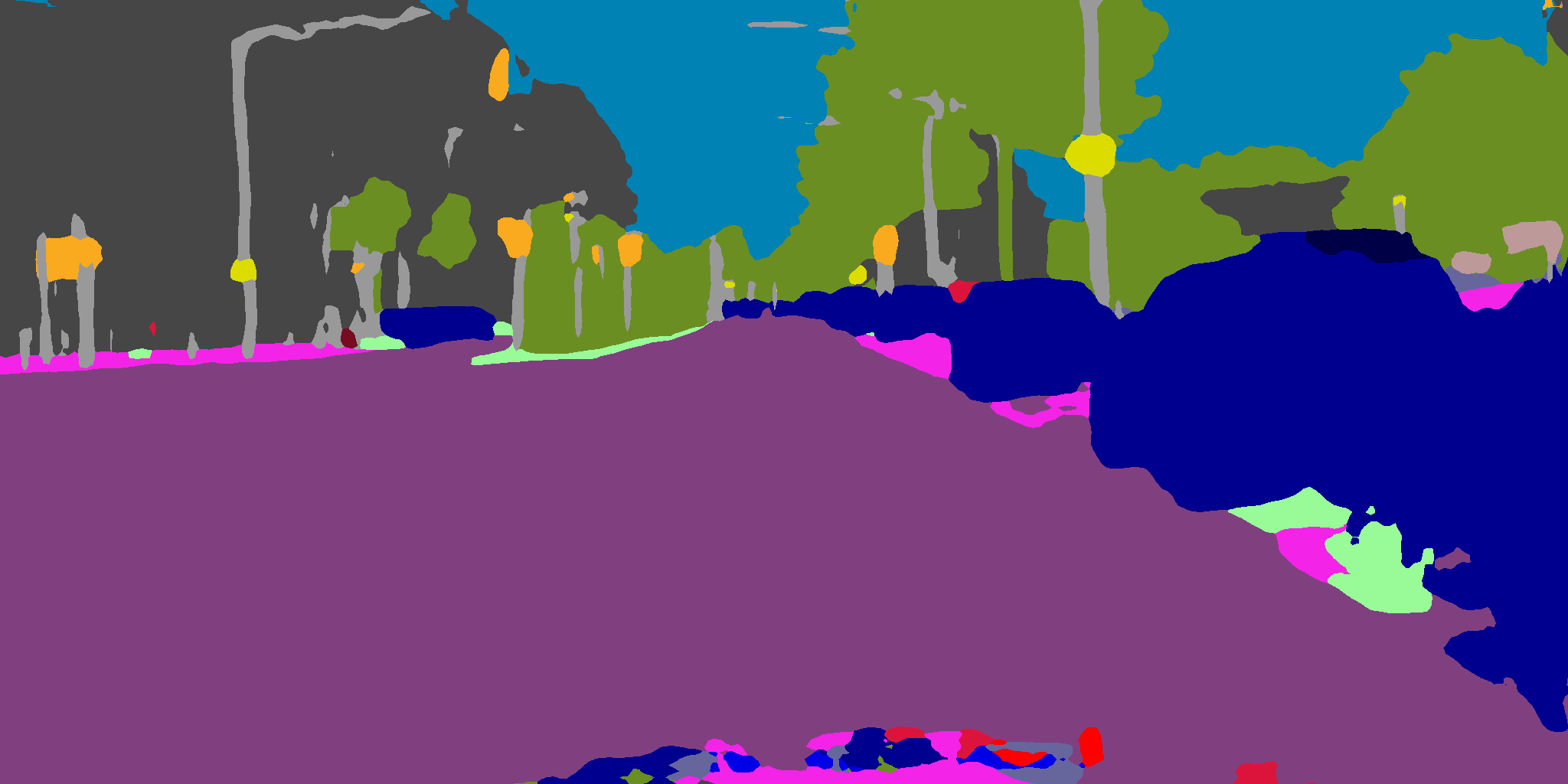} &
                \includegraphics[width=0.3\linewidth]{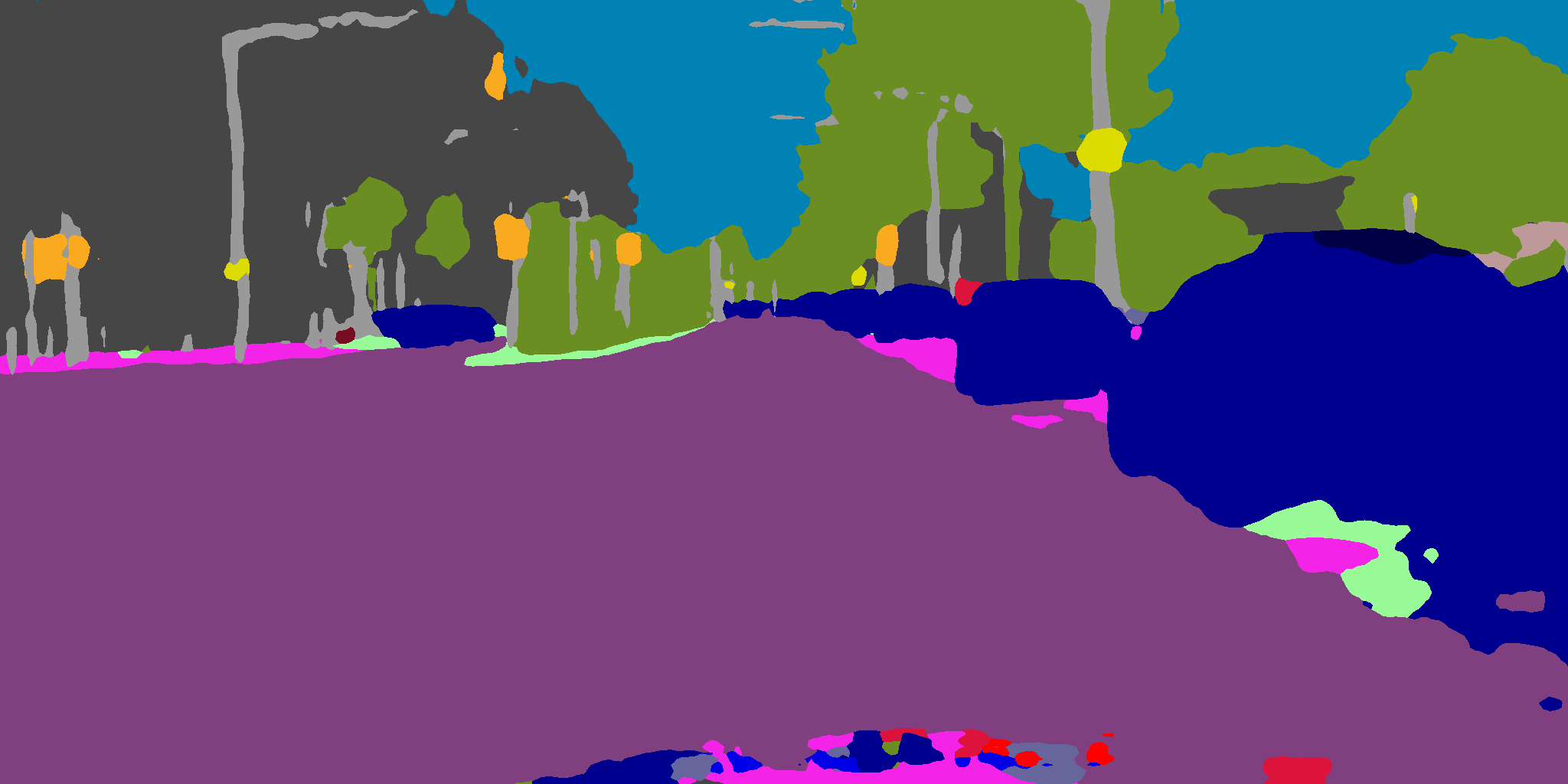} 

                \\
                
                \raisebox{1.5\normalbaselineskip}[0pt][0pt]{\rotatebox[origin=c]{90}{(c)}} &
                \includegraphics[width=0.3\linewidth]{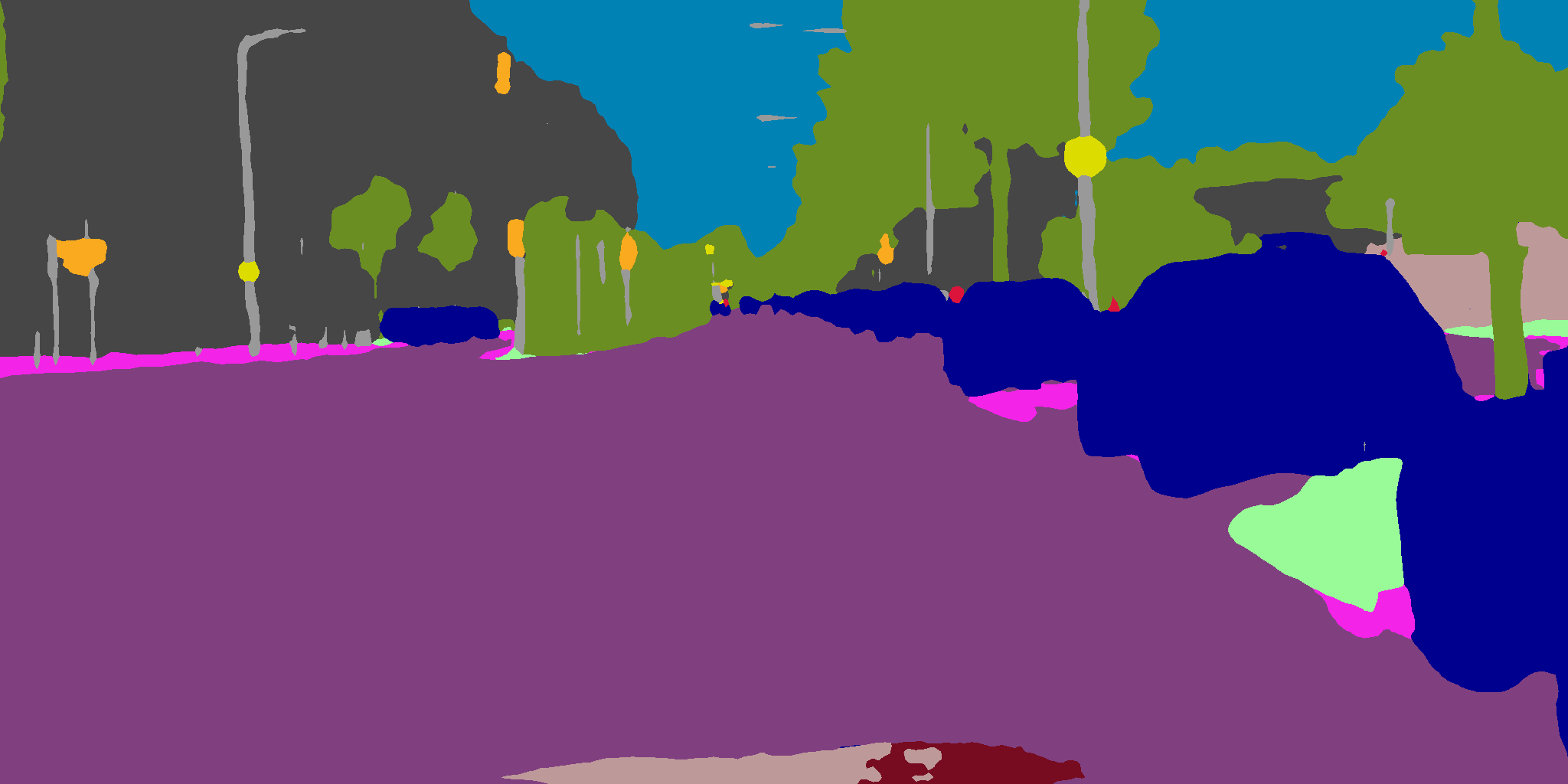} &
                \includegraphics[width=0.3\linewidth]{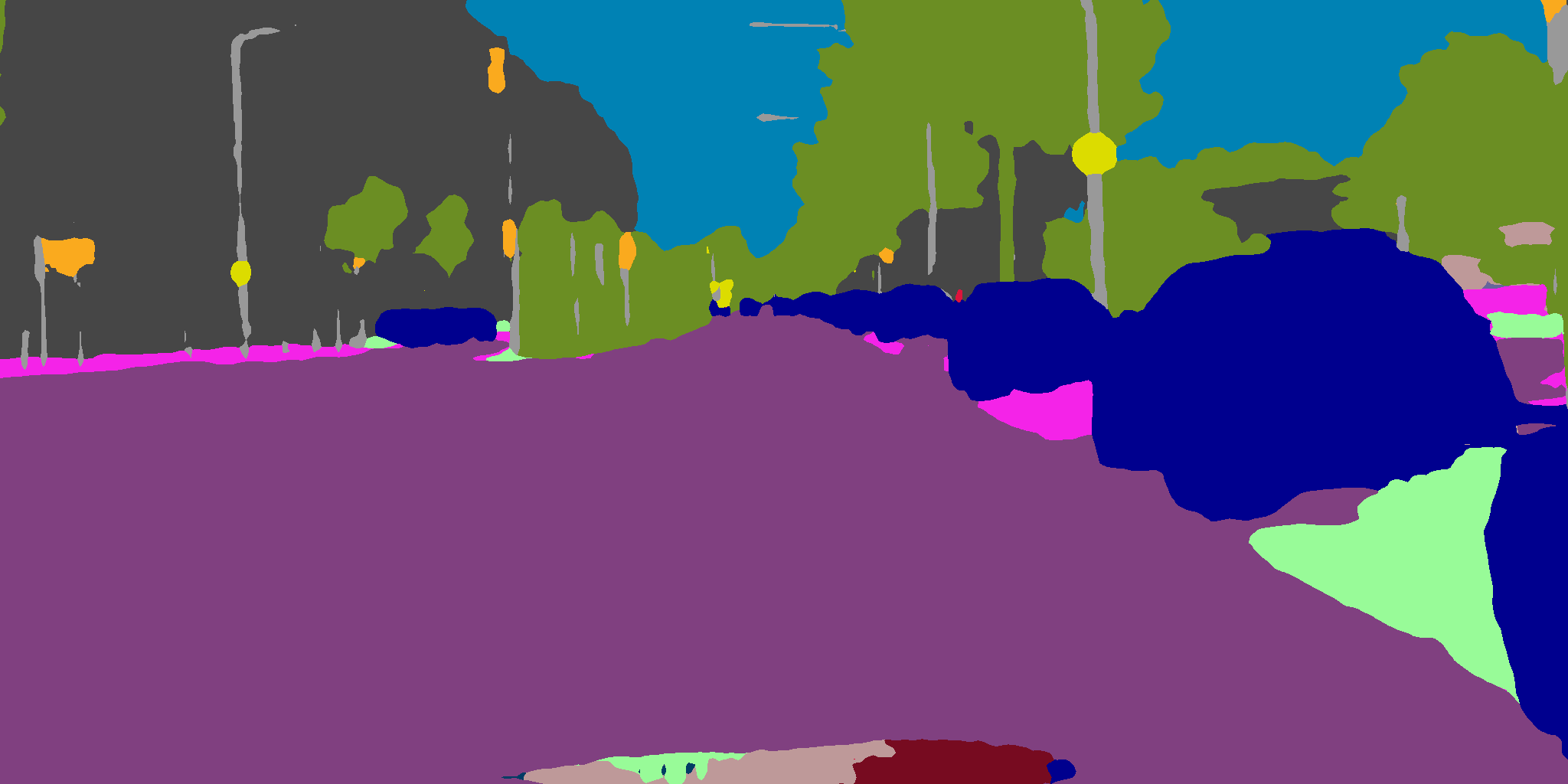} &
                \includegraphics[width=0.3\linewidth]{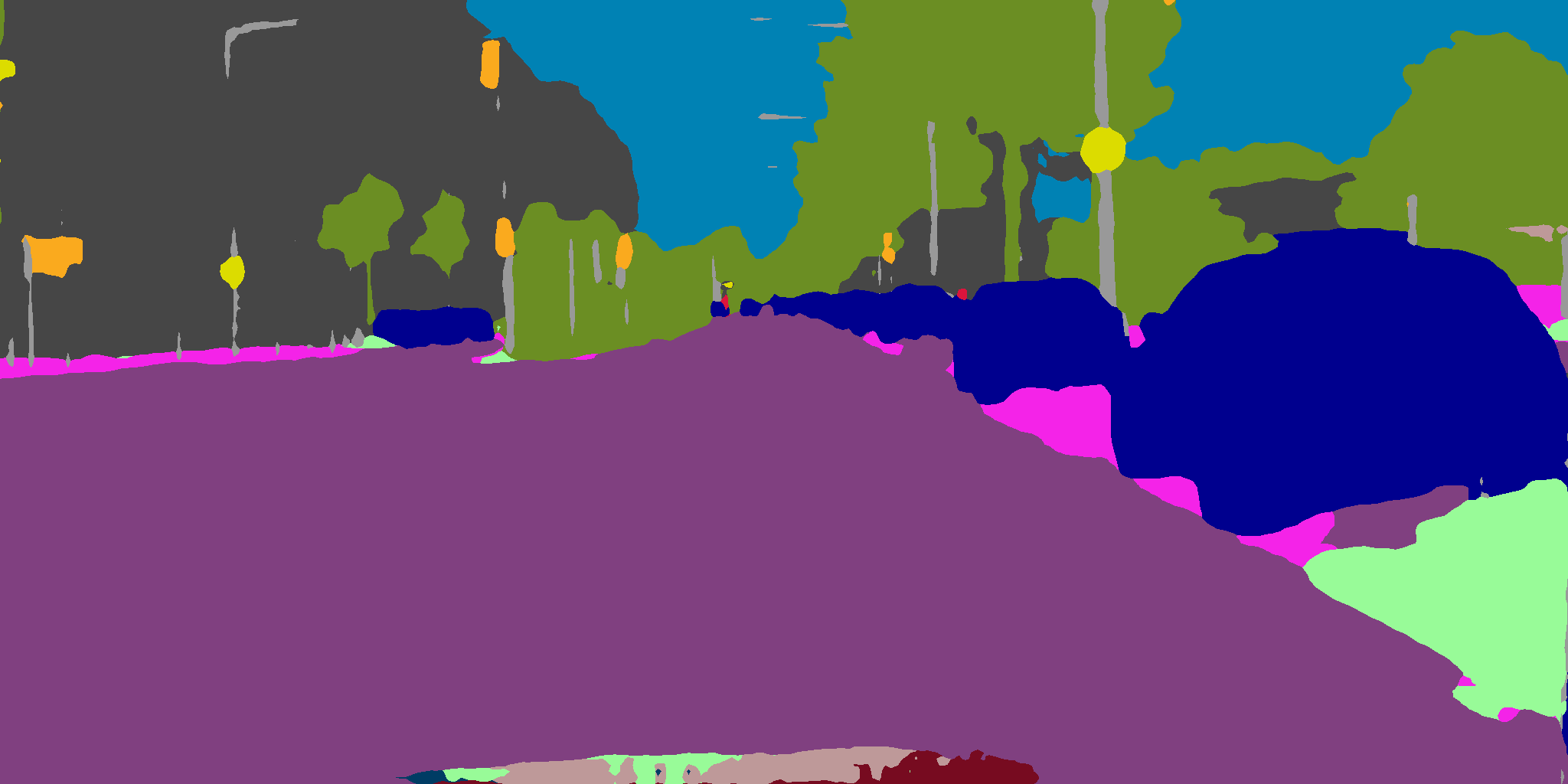}
                \\
                
                & k+1 & k+2 & k+3 \\    
            \end{tabular}
            \caption{\textcolor{black}{Qualitative evaluation: (a) Ours-SN-R18 (0.75), (b) A-Lite, and (c) SwiftNet-R18 (0.5). From left to right, each column with index k + n shows the segmentation results for a video frame that is n steps after the keyframe k. Note that only the last frame (k+3) has the ground-truth segmentation.
            }}
            \label{fig:qualitative_single_image}
            \vspace{-0.4cm}
        \end{figure}     
    
    \subsection{Ablation Experiments}\label{subsec:abs}
        This section aims at dissecting our design to understand better the contributions of each component. 
        
        \noindent \textbf{Using Lightweight Flow Estimation and Feature Extraction in A-R18:} \textcolor{black}{{Fig.~\ref{fig:ours_ax18_com} shows that Accel-R18, when trained end-to-end with our lightweight flow estimation network and feature extractor (a scheme termed A-Lite), suffers from a significant mIoU drop, although its throughput is much improved (red vs. cyan curves). This stresses the novelty of our guided dynamic filtering. Also shown in Fig.~\ref{fig:ours_ax18_com}, using FlowNet2s~\cite{flownet2s} in our framework has little impact on mIoU (blue vs. green curves). This however causes the throughput to reduce, due to the prolonged runtime for flow estimation.}}
        
        
        \noindent \textbf{Discarding Intra Features:} Intra features refer to the features extracted from the current frame. They serve the purposes of guiding the $1 \times 1 $ dynamic filtering and refining the semantic predictions propagated from the previous frame. Table~\ref{table:intra_com} shows discarding intra features causes the \textit{minimum} mIoU to drop much faster along with the increasing keyframe interval, highlighting their necessity for mitigating error propagation.  

    \subsection{Qualitative Evaluation}
        Fig.~\ref{fig:qualitative_single_image} presents qualitative evaluation. For a fair comparison, all the competing models are configured to have a similar throughput (120 to 140 FPS). Because of lightweight temporal propagation, video-based methods can afford  higher-accuracy keyframe segmentation, e.g. SwiftNet-R18 (0.75) in our method and BiSeNet-R18 (0.75) in A-Lite.y 
        We adopt A-Lite instead of A-X39 for comparison as it is the only version of Accel that achieves a similar throughput to the others. It is seen that our method produces consistently better segmentations across video frames (see the poles especially in the top-left area) than SwiftNet-R18 (0.5).
        In addition, A-Lite suffers from seriously distorted object boundaries. More qualitative results can be found in our supplementary document. 
\section{CONCLUSION}
\label{sec:conclusion}
This paper presents a simple propagation framework for efficient video segmentation. We show that it is more cost-effective to perform warping in segmentation output space than in feature space. This also allows the propagation error to be minimized at each time step by our guided spatially-varying convolution. Our scheme has the striking feature of being able to work with any off-the-shelf fast image segmentation network to further video segmentation.

\section{ACKNOWLEDGEMENT}
\label{sec:acknowledgement}
We are grateful to the National Center for High performance Computing for GPU resources. This project is supported by MOST 109-2634-F-009-007 through Pervasive AL Research (PAIR) Labs, National Yang Ming Chiao Tung University, Taiwan.

\bibliographystyle{IEEEbib}
\bibliography{icme2021template}

\end{document}